\documentclass[letterpaper, 10 pt, conference]{ieeeconf}  
\IEEEoverridecommandlockouts                              
\usepackage{epsfig} 
\usepackage{times} 
\usepackage{hyperref}
\usepackage{amsfonts}

\usepackage{amsmath} 
\usepackage{amssymb}  
\usepackage{mathrsfs}
\usepackage{comment} 
\usepackage{algorithm, algorithmic}

\title{\Large \bf
Sequential Motion Planning for Bipedal Somersault via Flywheel SLIP and Momentum Transmission with Task Space Control
}
    \author{Xiaobin Xiong and Aaron D. Ames
    \thanks{*This work is supported by NSF 1924526, NSF 1932091, NSF 1923239.}
\thanks{The authors are with the Department of Mechanical and Civil Engineering, California Institute of Technology, Pasadena, CA 91125
        {\tt\small xxiong@caltech.edu}, {\tt\small ames@caltech.edu}}%
 }
\begin{document}
\maketitle
\thispagestyle{empty}
\pagestyle{empty}

\begin{abstract}
In this paper, we present a sequential motion planning and control method for generating somersaults on bipedal robots. The somersault (backflip or frontflip) is considered as a coupling between an axile hopping motion and a rotational motion about the center of mass of the robot; these are encoded by a hopping Spring-loaded Inverted Pendulum (SLIP) model and the rotation of a Flywheel, respectively. We thus present the Flywheel SLIP model for generating the desired motion on the ground phase. In the flight phase, we present a momentum transmission method to adjust the orientation of the lower body based on the conservation of the centroidal momentum. The generated motion plans are realized on the full-dimensional robot via momentum-included task space control. Finally, the proposed method is implemented on a modified version of the bipedal robot Cassie in simulation wherein multiple somersault motions are generated.
\end{abstract}

\section{INTRODUCTION}
Human athletes perform somersaults as an ultimate demonstration of mobility and dynamic maneuverability. The anthropomorphic partner in the robotics world, the hydraulic-actuated robot Altas from Boston dynamics, was able to backflip \cite{atlas}, showing the decades of dedicated effort towards advancing robotic technologies; this dates back to the frontflips of the Raibert's hopper \cite{hodgins1988biped, playter1992control} in the 1980s. Regardless of the practical ramifications of doing somersaults on bipedal robots, achieving this kind of highly dynamic behavior benchmarks the stage of maturity of engineering robotic systems.

The back-flip of Atlas inspired the quadrupedal locomotion community which recently has built and controlled electric-powered quadrupedal robots to perform backflips \cite{minicheetah, aliengo}. However, these results have not been widely seen on full-scale bipedal robots, despite the existence of studies of bipedal somersaulting in \cite{tamada2014high, peng2018deepmimic, christiano2017deep}.

This is in part due to the fact that bipedal robots typically have more complex designs and, roughly speaking, less region of stability comparing to quadrupeds. These somersault behaviors require robot design to be able to make the behaviors feasible so that feedback control can play a role. The first observation is that the actuators need to be powerful to generate large vertical acceleration so that the robot can jump up to a certain height. The angular acceleration should be large enough so that enough momentum is generated on the ground to enable whole-body rotation is the flight phase.


Another observation is that backflipping requires the robot to have an upper body with large inertia and large ranges of motion at certain joints. To achieve the flip motion, the robot should have enough momentum at lift-off, which should be mainly generated by the upper body. This is because the lower body typically extends for jumping, the momentum of which cannot be freely generated. Additionally, the robot should have large ranges of motion at joints that connect to the linkages with large inertia; for instance, if the hip joints have small ranges of motion, then it requires impractical torques to accelerate the joints to the required rotational velocity within the ranges of motion.

\begin{figure}[t]
      \centering
      \includegraphics[width=3.3in]{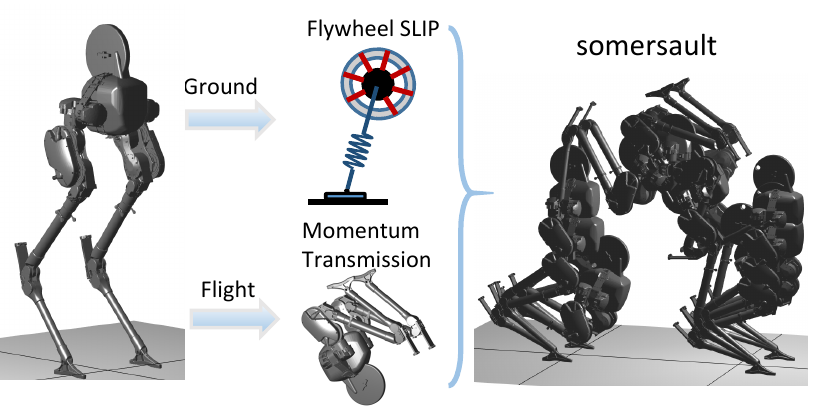}
      \caption{Illustration of the proposed methodology for generating somersaults.}
      \label{overview}
\end{figure}

As illustrated thus far, achieving somersaults is a complex system-level problem of design and control. In this paper, we mainly investigate the motion planning and feedback control component in achieving these motions. The robot design to make somersaults feasible is out of the scope of this paper, even though it is an important and challenging problem. 

In \cite{atlasDW, minicheetah}, the motion plans for backflips are generated via trajectory optimization on the full-dimensional model of the quadrupeds, which has computational challenges for more complex bipedal robots. The formulated optimization is also highly nonconvex in a large dimensional space \cite{hereid20163d, posa2014direct, ding2018single}. The optimization typically requires good initial guesses to solve, which can have feasibility issues. If the robot design renders the motion infeasible to realize, the trajectory optimization runs into infeasibility and provides little information towards the possible improvement in the robot design. Moreover, the optimized plans based on the full model typically are not robust to modeling errors and external disturbances. Thus a principled simple representation, e.g. the Linear Inverted Pendulum (LIP) \cite{kajita20013d} for walking and the Spring-loaded Inverted Pendulum (SLIP) \cite{altendorfer2001evidence} for running, is preferred for studying the essential components which realize the somersault.



To address this, we present a Flywheel SLIP (FSLIP) model to capture somersault dynamics on the ground phase, including jumping and landing phases. The FSLIP has a flywheel \cite{pratt2006capture, stephens2011push, takenaka2009real} which is attached on a SLIP \cite{xiong2018coupling} model. The point mass dynamics of the FSLIP represents the hopping part of the somersault, while the flywheel captures the centroidal angular dynamics \cite{nava2017momentum}. Trajectory optimization is formulated on the FSLIP to generate jumping and landing behaviors subject to the task constraints and additional physical constraints of the somersault. With the low dimensionality, it is easy and fast to solve the optimization.

When the robot is in the flight phase, the centroidal angular momentum \cite{orin2013centroidal, saccon2017centroidal, lee2007reaction, koolen2016design, hofmann2009exploiting} is conserved. We present a moment transmission control for a specific linkage to modulate the angle of the rest of the body. The specific linkage can be a flywheel, a tail, or arms of the robot. Moreover, additional assistive strategies are presented towards the landing.

The motion planning from the reduced-order representation is generated in sequence to the domains of the somersault. The motion plan is realized on the full-dimensional robot via an optimization-based controller. The optimization-based controller also takes into consideration the ground reaction force constraints and the input constraints.

The proposed method is realized on a modified version of the bipedal robot Cassie in simulation. We added a flywheel on the top of its pelvis to increase the inertia of its upper body and to amplify its capability to control the momentum. The flywheel can also be viewed as a representation of arms on the robot that have a free range of motion at the shoulders. Various somersault motions are generated, including front/back-flips and frontflips with over 1m forward jumping distance. Robustness and practical feasibility are also analyzed and discussed. With the formulated FSLIP model and the correspondence to the realization on a full dimensional robot, we hope that this study can shed some light on bipedal robot design and motion generation towards realizing versatile aerial motion on bipedal robots. 

\section{Modelling}
\label{sec:Modeling}
We start by presenting the model of the robot with the modification. Then we model the hybrid dynamics of the somersault, which is used in the simulation and analysis of the proposed planning and control methods.

\subsection{Robot Model}
The robot Cassie is a bipedal robot that and its predecessor ATRIAS \cite{grimes2012design} are built to best mimic the canonical Spring-loaded Inverted Pendulum (SLIP) model. As a result, it has a concentrated upper body and mechanical springs in the leg. The robot has shown good mechanical properties for walking \cite{xie2018feedback, gong2018feedback}. However, the trivial torso on Cassie makes it extremely difficult to produce any movement with aerial phases. 

\begin{figure}[t]
      \centering
      \includegraphics[width=3in]{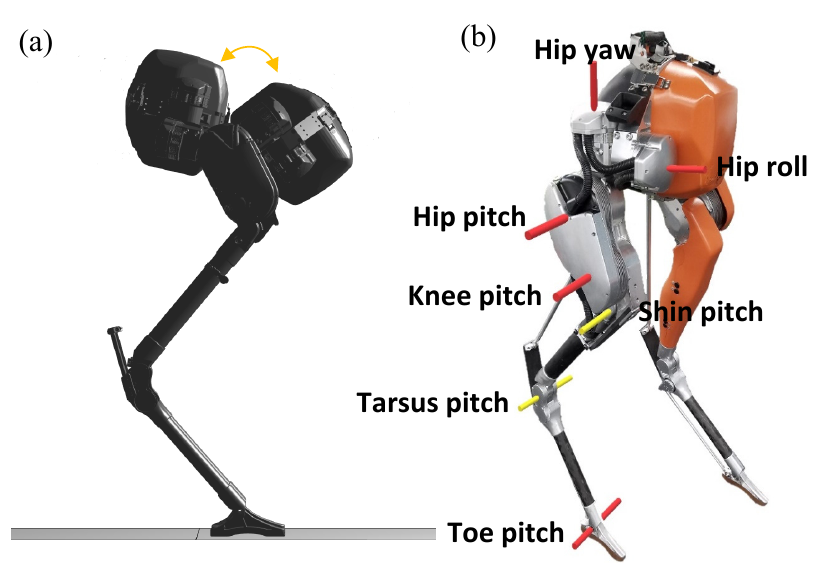}
      \caption{(a) The range of motion of the pelvis. (b) The joints on the lower body of the bipedal robot Cassie by Agility Robotics.}
      \label{robot}
\end{figure}

For instance, in \cite{XiongSLIP}, vertical hopping is realized on the robot with a goal of producing zero momentum before the flight phase. The leg extension generates a non-zero angular momentum, and thus the torso has to cancel out the angular momentum by producing a large forward rotation of the pelvis, which is limited by its range of motion (see Fig. \ref{robot} (a)), and maximum speed at the hip pitch joint. In other words, with a small torso to generate a large momentum, the required torso rotational speed is impractical to be realized within the range of motion.  
Thus using the trivial torso to generate somersaults also becomes impractical. To enable the capability, we present the following modification.

\textbf{Hardware Modification:} A flywheel is added on the top of the pelvis of the robot Cassie. It can be viewed as a robotic tail in some sense, e.g. the tail on Salto \cite{haldane2017repetitive}. The design can be optimized for the location and size of the flywheel. Since the focus of this paper is on the planning and control, we manually select the installation of the flywheel based on intuition. The flywheel is installed 15cm above the pelvis, assuming the flywheel is made of aluminum. The exact specifications of the flywheel are discussed in section \ref{sec:results}. 
Except the flywheel, all physical parameters of the robot remain unchanged.

\textbf{Dynamics Model:}
We use a model of 22 degrees of freedom (DOFs) to represent the robot Cassie \cite{XiongSLIP}. With the flywheel modification, the robot has 23 DOFs. Let $q \in SE(3) \times \mathbb{R}^{17}$ represent the configuration of the robot. The Euler-Lagrange equation for its dynamics is:
\begin{eqnarray}
&& M(q)\ddot{q} + H(q,\dot{q}) = Bu+J_s^T \tau_s + J_{h,v}^T F_{h,v},  \label{eom} \\
&& J_{h,v}(q) \ddot{q} + \dot{J}_{h,v}(q) \dot{q} = 0, \label{hol}
\end{eqnarray}
where $M(q)$ is the mass matrix, $H(q,\dot{q})$ is the Coriolis, centrifugal and gravitational term, $B$ and $u\in \mathbb{R}^{11}$ are the actuation matrix and the motor torque vector, $\tau_s$ and $J_s$ are the spring joint moment vector and the corresponding Jacobian, and $F_{h,v}\in \mathbb{R}^{n_{h,v}}$ and $J_{h,v}$ are the holonomic force vector and the corresponding Jacobian respectively. The subscript $v$ is used to indicate different domains which have different numbers of holonomic constraints.

\subsection{Hybrid Model of Somersaulting}
The somersault is a hybrid dynamical phenomenon with ground and flight phases. We assume that the robot hops in the sagittal plane and its left and right feet have the same contact mode. For simplification, we also assume the feet always flatly contact the ground, which will be encoded as one of the control objectives. Based on the sequence of the contact modes, we divide the motion into three domains, i.e., $v =$ jumping, flight, and landing. In jumping, the feet are in contact with the ground. It transits from jumping to flight when the ground reaction normal forces reach to zero. In the flight phase, the feet are above the ground. The robot goes through a large whole-body rotation in its sagittal plane. It transits into landing when the feet strike the ground. The foot-ground impact is modeled as plastic impact \cite{grizzle2014models}, where the position trajectories are smooth while the velocities are discontinuous.

\section{The Flywheel SLIP Model}\label{SLIPmodel}
In this section, we explain the Flywheel SLIP (FSLIP) model for generating the jumping and landing trajectories for the robot. The point mass dynamics of the FSLIP captures the hopping dynamics of the center of mass (COM) dynamics of the robot; the rotational dynamics of the flywheel corresponds to the rotational part of the centroidal dynamics of the robot. Using parametric regressions on the stiffness of the leg spring and the centroidal inertia of the robot, we identify the Flywheel SLIP model to best capture the features of the robot.

\subsection{The Dynamics}
\begin{figure}[t]
      \centering
      \includegraphics[width = 1\columnwidth]{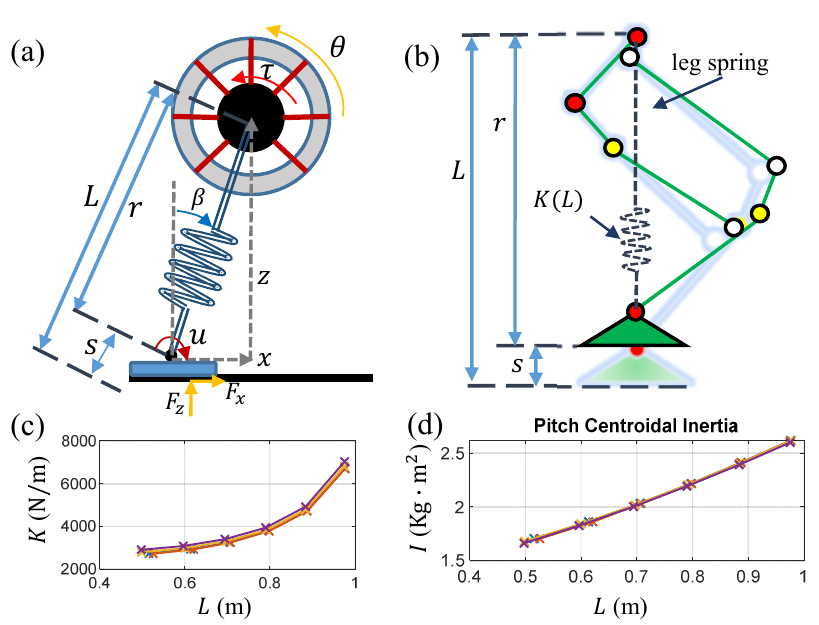}
      \caption{(a) The Flywheel SLIP. (b) Illustration of the leg spring from the leg of Cassie. (c, d) The stiffness of the leg spring and the pitch centroidal inertia of the robot v.s. the leg length for different standing configurations.}
      \label{SLIP}
\end{figure}

The Flywheel SLIP is the actuated SLIP model in \cite{xiong2018coupling, xiong2019IrosStepping} with a flywheel attached on the mass and an actuated foot (see Fig. \ref{SLIP} (a)). The actuated SLIP is the canonical SLIP model with leg length actuation and a prismatic spring. The flywheel is actuated to rotate about the mass location. The foot actuation is added to work with the flywheel to generate appropriate acceleration. The dynamics are given by:
\begin{align}
m \ddot{x} &=-\frac{\tau + u}{r}\mathrm{cos}(\beta) + F^s \mathrm{sin}(\beta) \\
m \ddot{z} &= - m g  + \frac{\tau +u}{r}\mathrm{sin}(\beta) + F^s \mathrm{cos}(\beta) \\
 I \ddot{\theta} &= \tau  \\
\ddot{s} &= \ddot{L} - \ddot{r}
\end{align}
where:
\begin{itemize}
    \item $m$ is the mass of the mass and the flywheel,
    \item $I$ is the rotational inertia of the flywheel,
    \item $\tau$ is the torque on the flywheel,
    \item $u$ is the foot actuation,
    \item $\theta$ is the angle of the flywheel,
    \item $\beta$ is the angle of the leg,
    \item $r$ is the real leg length,
    \item $L$ is the actuated leg length,
    \item $s$ is the spring deformation,
  \item $F^s$ is the spring force,
    \item $x, z$ are the horizontal and vertical mass positions. 
\end{itemize} The kinematic constraints are $s + r = L$ and
$\dot{s} + \dot{r} = \dot{L}$. The spring force is calculated by: $F^s = Ks + D \dot{s}$,
where $K, D$ are the stiffness and damping of the spring.

\textbf{Correspondence to the robot:}
The Flywheel SLIP model described above is general. Here we use parametric regressions to best match the model to the robot. Similar to \cite{XiongSLIP, xiong2018coupling}, we identify the stiffness and damping of the leg spring from the joint springs of the robot. Then parametric regressions are used to project the stiffness $K$ and damping $D$ of the leg spring over the leg length $L$. Thus the leg spring of the FSLIP mimics the leg of the robot, as illustrated by Fig. \ref{SLIP} (b). Additionally, we regress the centroidal pitch inertia \cite{orin2008centroidal} $I(L)$ of the robot, which becomes the inertia of the Flywheel. It is important to note that it is the inertia of the whole robot instead of the flywheel on the pelvis that corresponds to the flywheel on the FSLIP. The regressed FSLIP model from the robot is expected to produce behaviors which can be translated to Cassie. Fig. \ref{SLIP} (c) (d) show the leg stiffness of the leg spring and the pitch inertia over different leg length for different configurations, indicating that they are mainly changed by $L$.

\subsection{Jumping Optimization}
Now we use the Flywheel SLIP model to generate a jumping motion from standing for enabling the somersault. The planned motion is required to satisfy the actuation constraints, the ground force constraints, the required flipping momentum, and the apex constraints. The planned motion contains the trajectories of the point mass and desired centroidal momentum, which will be tracked by the robot.

Similar to \cite{XiongSLIP}, we use a direct transcription method \cite{betts1998survey, hereid20163d} to describe and solve the optimization problem. The trajectories are discretized and the dynamics are integrated via trapezoidal integration. Here we mainly describe the constraints and the cost function of the optimization.

\textbf{Task Constraints:}
The goal of the jumping optimization is to create enough vertical velocity and angular momentum to enable somersaults. The task is encoded via the velocities at the lift-off event. We assume that the robot lands with a similar self-configuration to that at lift-off. Thus the duration of the flight phase is given by:
\begin{equation}
T_{\text{Flight}} = \frac{2 \dot{z}^{\text{lift-off}}}{g},
\end{equation}
where $\dot{z}^{\text{lift-off}}$ is the vertical velocity of the point mass at lift-off. If the robot fixes its configuration during the flight phase, the angular velocity of the whole body at lift-off should be:
\begin{equation}
\label{taskEncoding}
\dot{\theta}^{\text{lift-off}} = \frac{\pm2 \pi - 2 \beta^{\text{lift-off}}}{T_{\text{Flight}}} =  \frac{ (\pm\pi  -\beta^{\text{lift-off}}) g}{\dot{z}^{\text{lift-off}}}.
\end{equation}
Note that we will relax the condition of the fixed configuration on the robot later. Eq. \eqref{taskEncoding} establishes the constraint on the jumping phase to produce a flip motion. The sign $\pm$ depends on if the somersault is a backflip or frontflip.

\textbf{Physical Constraints:} During jumping, we enforce non-slipping condition on the foot via,
\begin{equation}
\quad \quad   \quad \quad  \quad \quad    -F_z \mu \leq F_x \leq F_z \mu,  \nonumber  \quad \quad\quad \quad \text{(Friction Cone)}
\end{equation}
where $F_z$ is the vertical ground reaction force (GRF), and $\mu$ is the friction coefficient. It is also desirable to keep the feet flat on the ground, thus the zero moment point (ZMP) constraint is included:
\begin{equation}
\quad \quad\quad  \quad \quad\quad   \quad \  - F_z \frac{l}{2} \leq u \leq F_z \frac{l}{2},  \nonumber  \quad \quad\quad \quad \quad\quad \text{(ZMP)}
\end{equation}
where $l$ is the foot length. Additional constraints include the actuation limits, leg length limits and non-negative normal ground reaction forces. The final ground normal force should reach to 0 to trigger the lift-off event, i.e., $F_z^{\text{lift-off}}=0$.

\begin{figure}[t]
      \centering
      \includegraphics[width= 1\columnwidth]{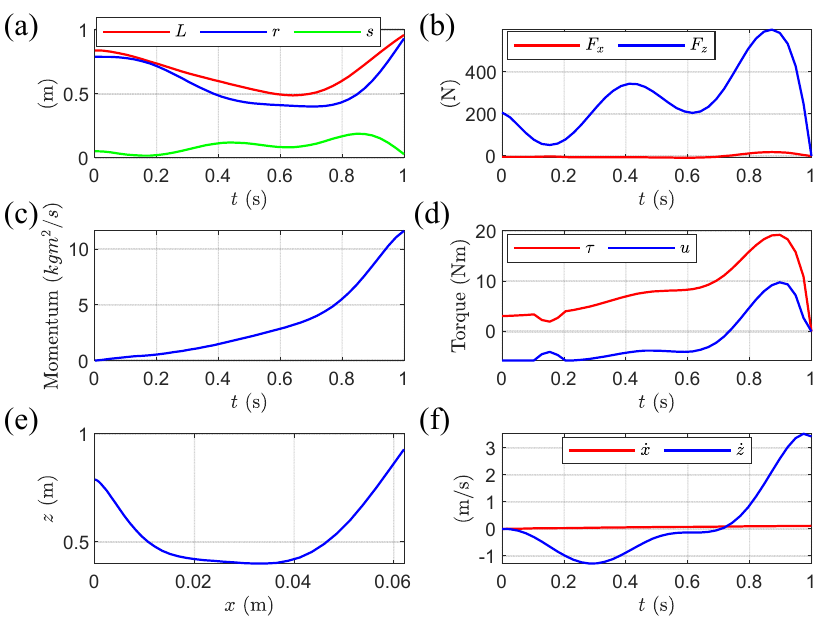}
      \caption{An example of the optimized jumping trajectories of the leg length and spring deformation (a), ground reaction forces (b), the momentum of the flywheel (c), the torques (d), the position of the point mass (e) and the velocities of the mass (f). }
      \label{jumpingOpt}
\end{figure}

\textbf{Cost:} The cost for the jumping optimization is on the virtually consumed energy:
\begin{equation}
J_{\textrm{Jumping}} = \textstyle \sum_{i} (c_1 \ddot{L}_i^2 +  c_2 u_i^2 +  c_3 \tau_i^2) \Delta T,
\end{equation}
where $c_{1,2,3}$ are the coefficients, $\Delta T$ is time discretization, and $i$ is the index for the discretized variables. Fig. \ref{jumpingOpt} illustrates an example of the optimized trajectories of a jumping behavior.



\subsection{Landing Optimization}
The landing optimization is performed after the touch-down event at the end of the flight phase. Since it uses the same Flywheel SLIP model, we briefly explain the landing planning here.

During landing, the system rests to a static configuration from the flight phase. The task can be directly encoded via the equality constraints on the final states:
\begin{align}
z_N &= z^{\text{des}}, \nonumber\\
\dot{z}_N  &= x_N = \dot{x}_N = \dot{\theta} = 0,\nonumber \\
F^s_N  &= mg,\nonumber
\end{align}
where the subscript $N$ indicates the final state of landing. The physical constraints are the same to that of the jumping planning. The initial states are constrained based on the post-impact state of the robot. The cost is the cost of jumping plus a penalty on the spring oscillation:
\begin{equation}
J_{\textrm{Landing}} =J_{\textrm{Jumping}} + \textstyle \sum_{i} \alpha \dot{s}_i ^2  \Delta T ,
\end{equation}
where $\alpha$ is a weighting coefficient.

\section{Flight Phase Planning}

In this section, we present the planning methodology for the flight phase. The core idea here is to change the momentum of the flywheel to control the configuration of the lower body of the robot. This is due to the conservation of the angular centroidal momentum of the robot in the flight phase. Additionally, we add two components to assist the flipping motion with a view toward landing the jump.

\subsection{Momentum Transmission on the Flywheel}
As the flight phase is underactuated due to the conservation of momentum, direct controlling the rotation of the floating robot is not possible. However, it is possible to change the momentum of the lower body of the robot (linkages except the flywheel) by controlling the momentum of the flywheel. The total centroidal pitch momentum $\mathcal{H}_{\text{pitch}}$ of the robot in the flight phase is,
\begin{equation}
\mathcal{H}_{\text{pitch}} = I_{\text{flywheel}} \omega_{\text{flywheel}}  + \textstyle \sum_{i = 2}^N I_{i} \omega_i,
\end{equation}
where $I_{\text{flywheel}}, \omega_{\text{flywheel}}$ are the pitch inertia and velocity of the flywheel about the COM of the robot, and $I_{i}, \omega_i$ are the pitch inertia and velocity of the other linkages about the COM. The conservation of momentum indicates $\mathcal{H}_{\text{pitch}}$ is constant. The averaged angular velocity of lower body can be defined as: \begin{equation}
\bar{\omega} = \frac{\textstyle \sum_{i = 2}^N I_{i}\omega_i}{\textstyle \sum_{i = 2}^N  I_{i}},
\end{equation}
which is controllable by changing the momentum of the flywheel. If we assume the robot is rigid, the flipping condition becomes,
\begin{equation}
\textstyle \int_0^{T_{\text{Flight}}} \bar{\omega} (t) = \pm 2\pi - 2 \beta^{\text{lift-off}}.
\end{equation}
The initial pitch angle of the lower body is $\theta(t=0) = \beta^{\text{lift-off}}$, and the final angle at touch-down is $\theta(t = T_{\text{Flight}}) = \pm 2\pi - \beta^{\text{lift-off}}$, where $\pm$ depends on the direction of the somersault. The initial velocity $\bar{\omega}^{\text{lift-off}}$ is calculated from the robot, and we let the final velocity $\bar{\omega}^{\text{touch-down}} = 0$. Thus, trajectories of $[\theta^{\textrm{des}}(t), \bar{\omega}^{\textrm{des}}(t)]$ can be constructed based on the boundary conditions. Since it is desirable to adjust the pitch angle of the lower body to prepare for landing at good configurations, we add a feedback law on the desired average velocity:
\begin{equation}
\dot{\theta}_{cl}^{\textrm{des}}(t) = \bar{\omega}^{\textrm{des}}(t) - k_p (\theta_{\textrm{pelvis}}(t)  - \theta^{\textrm{des}}(t)),
\end{equation}
where we will use the pelvis pitch angle $\theta_{\textrm{pelvis}}$ as a reference for the orientation of the lower body. Thus we control the pelvis pitch angle to be the same of the leg angle $\beta$ on the ground phase. Finally, the desired centroidal momentum of the flywheel is:
\begin{equation}
\label{momentumTrans}
\mathcal{H}^{\textrm{des}}_{\text{flywheel}}(t) = \mathcal{H}_{\text{pitch}} - \dot{\theta}_{cl}^{\textrm{des}}(t) (\textstyle \sum_{i = 2}^N I_{i}).
\end{equation}

The rotational momentum required to achieve the whole body rotation is mainly accumulated on the flywheel during the jumping phase. In the flight phase, the momentum is transmitted from the flywheel to the lower body to achieve the desired rotation. Thus we refer Eq. \eqref{momentumTrans} as the \textit{momentum transmission} of the flywheel.

\subsection{Assistive Components}
In addition to the control of the flywheel, the lower body is also controlled to assist the somersault. We consider two assistive components as follows:

\textbf{Leg Retraction:} The legs of the robot can retract in the beginning of the flight phase and then extend towards the landing. This is referred to as \textit{tucking} in \cite{hodgins1988biped}. The purpose of the tucking is to decrease the centroidal inertia and thus increase the angular rotational rate. We implement this strategy by controlling the leg length. The desired trajectory is constructed smoothly with appropriate timing.

\textbf{Foot Placement:} The robot is expected to achieve the desired angular rotation and then land on the ground. To have the appropriate landing configuration, the robot should control the horizontal positions of the feet, which is controlled by adjusting the leg angle to the opposite of the leg angle at lift-off. By symmetry, the opposite leg angle can ensure the feasibility of landing in the optimization. Additionally, we control the feet to be flat to prepare for landing.



\section{Output Construction and Stabilization}
In this section, we explain the control of the aforementioned plans as they are realized on the 3D robot Cassie. We first explain the output selection and construction in each domain, then derive the optimization-based control method for stabilization.

In our previous work \cite{XiongSLIP, westenbroek2019}, we applied the control Lyapunov function based quadratic programs (CLF-QPs) \cite{ames2014rapidly} for output stabilization with momentum outputs. A different optimization-based control, i.e., task space control (TSC) \cite{xian2004task, wensing2013generation, escande2014hierarchical, XiongRal2020}, can serve the same purpose of stabilization. Here we derive TSC with minimum variable formulation for illustration.

\subsection{Output Construction}
The output construction is used to specify the desired behavior from the Flywheel SLIP model and the momentum transmission for the robot via featured functions (outputs) of the states. There are different outputs to be specified in the ground and flight phases.

\begin{figure*}[t]
      \centering
      \includegraphics[width= 6.6in]{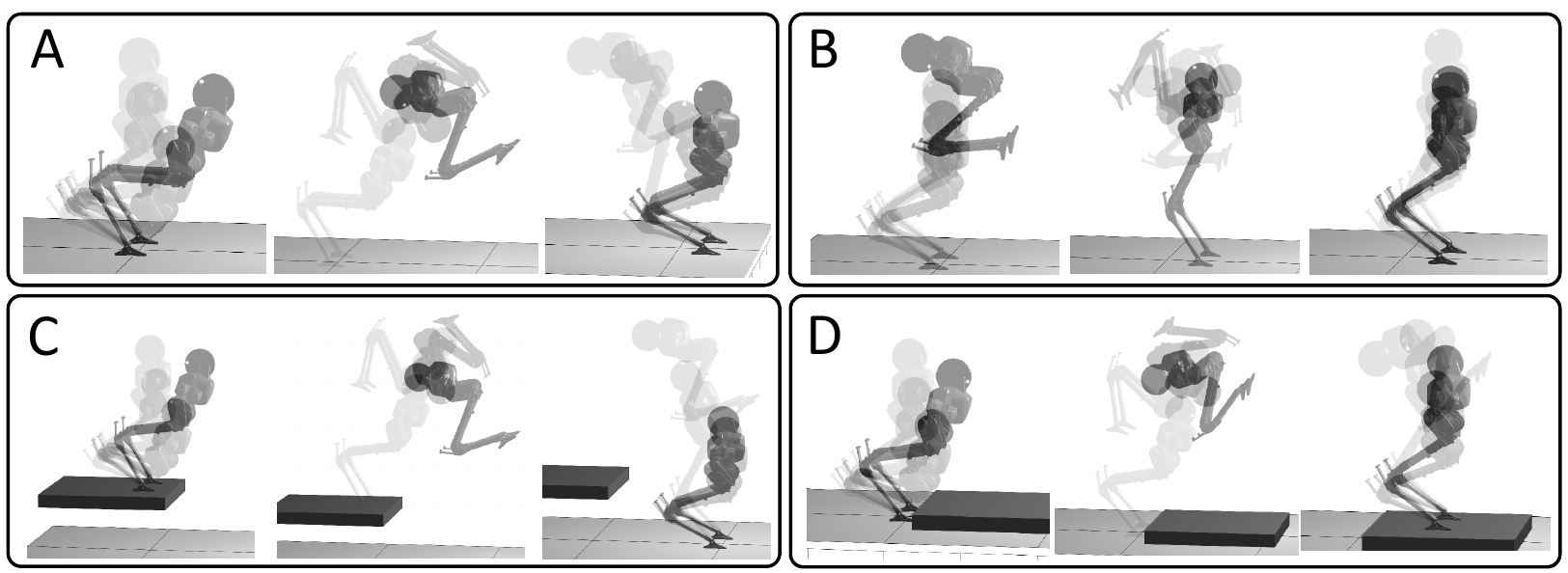}
      \caption{Snapshots of the generated somersaults: (A) a frontflip with a 1m forward target distance, (B) a backflip, (C) a frontflip from a platform  of $0.5$m height, and (D) a frontflip without knowing the landing surface, which is with a $0.1$m height from the ground.}
      \label{results}
\end{figure*}
\begin{figure}[!b]
      \centering
      \includegraphics[width= 3.3in]{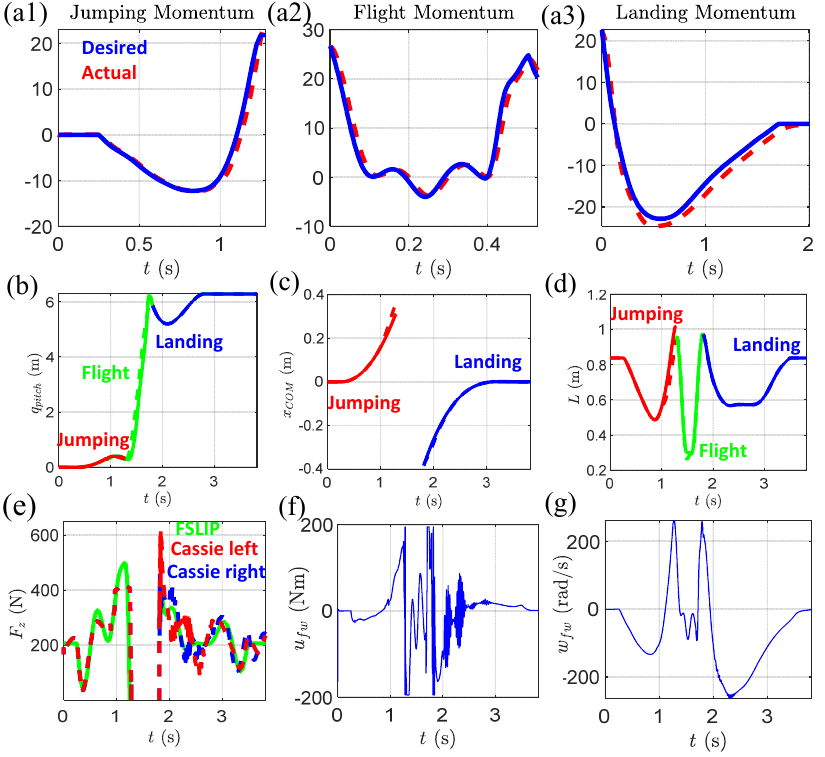}
      \caption{Simulation plots of the forward frontflip A (dashed lines are the actual; solid lines are the desired): (a1-a3) The momentum output trajectories in each domain (The flywheel momentum is the output in (a2)). (b) The leg length trajectories. (c) The ground norm forces of Cassie v.s. that of the FSLIP. (d) $x_\text{COM}$ (e) The pitch angle of the pelvis. (f) The applied torque on the flywheel. (g) The rotational speed of the flywheel. }
      \label{results2}
\end{figure}
\subsubsection{Jumping/Landing} The Flywheel SLIP provides the desired point mass dynamics and rotational dynamics for the robot. The point mass dynamics in the leg axial direction is enforced by the leg length $L_{\textrm{robot}}(q)$ on the robot; the mass dynamics in the translational/forward direction is enforced by the COM forward position $x_{\textrm{com}}(q)$ of the robot.
The rotational dynamics is enforced via the centroidal pitch momentum $\mathcal{H}_{\textrm{Pitch}}(q, \dot{q})$ of the robot. Additionally, we enforce zero motion in the coronal and transversal planes by constraining the lateral position of the COM $y_{\textrm{com}}(q)\rightarrow 0$ and the yaw angle of the pelvis $\phi_{\textrm{yaw}}(q) \rightarrow 0$. The outputs are the same for the jumping and landing phase, which are:
\begin{eqnarray}
&& \mathcal{Y}_1^{J/L}(q, \dot{q}, t)=\mathcal{H}_{\textrm{Pitch}}(q, \dot{q}) -  \mathcal{H}_{\textrm{FSLIP}}(t),\\
&& \mathcal{Y}_2^{J/L}(q, t) = \begin{bmatrix} L_{\textrm{robot}}(q) \\ x_{\textrm{com}}(q) \\ y_{\textrm{com}}(q) \\\phi_{\textrm{pitch}}(q) \\ \phi_{\textrm{yaw}}(q) \end{bmatrix}   -
 \begin{bmatrix} L_{\textrm{FSLIP}}(t) \\ x_{\textrm{FSLIP}}(t)\\ 0\\ \phi^{\textrm{des}}_{\textrm{pitch}}(t) \\0 \end{bmatrix}.
\end{eqnarray}
where $\mathcal{H}_{\textrm{FSLIP}}(t) = I_{\textrm{FSLIP}}(t) \theta (t) $ is the momentum of the Flywheel SLIP, $x_{\textrm{FSLIP}}(t)$ is the forward position of the Flywheel SLIP, and $\phi^{\textrm{des}}_{\textrm{pitch}}(t)$ is the desired pitch angle of the pelvis by construction. The subscript $_{1/2}$ indicates the number of times one must differentiate to have the torque $u$ appear.

\subsubsection{Flight}
As mentioned in the previous section, the flight phase is underactuated. The pitch momentum is transmitted between the flywheel and the lower body by Eq. \eqref{momentumTrans}. Thus the momentum of the flywheel $\mathcal{H}_{\textrm{flywheel}}(q, \dot{q})$ is controlled:
 \begin{equation}
\mathcal{Y}_1^{F}(q, \dot{q})= \mathcal{H}_{\textrm{flywheel}}(q, \dot{q}) -  \mathcal{H}^{\textrm{des}}_{\textrm{flywheel}}(t).
\end{equation}
To enable body tucking, the leg length $L_{\textrm{robot}}(q)$ are controlled to retract first and then extend. The rest joints of the body $ q_{/\textrm{knee}}$, i.e. motor joints except the knee joints, are controlled as fixed. Thus the outputs are defined as:
\begin{eqnarray}
 \mathcal{Y}_2^{F}(q, t) = \begin{bmatrix}  L_{\textrm{robot}}(q) \\ q_{/\textrm{knee}}  \end{bmatrix}   -
 \begin{bmatrix} L^{\textrm{des}}(t)\\ q^{+}_{/\textrm{knee}} \end{bmatrix}, \ \textrm{if} \  t < T_1
\end{eqnarray}
where $T_1$ is manually decided, e.g., $T_1 = \frac{3}{4} T_\textrm{Flight}$.
As approaching landing, it is desirable to prepare for the touch-down event by controlling the leg angle $\beta_{\textrm{leg}}(q)$ and toe pitch angle $q^{\textrm{pitch}}_{\textrm{toe}}(q)$. The outputs are:
\begin{eqnarray}
 \mathcal{Y}_2^{F}(q, t) = \begin{bmatrix}  L_{\textrm{robot}}(q) \\  \beta_{\textrm{leg}}(q) \\ q^{\textrm{pitch}}_{\textrm{toe}}(q)   \\ q_{\textrm{rest}}  \end{bmatrix}   -
 \begin{bmatrix} L^{\textrm{des}}(t)\\ -\beta^{+}_{\textrm{leg}} \\ 0\\ q^{+}_{\textrm{rest}} \end{bmatrix},\  \textrm{if} \  t \geq T_1,
\end{eqnarray}
where $\beta^{+}_{\textrm{leg}}$ is the leg angle at the lift-off event. $q_{\textrm{rest}}$ denotes the rest motor joints except for the constrained one, which in this case are the hip roll and yaw joints.
\subsection{Momentum-Included Task Space Control}

Finally we present the output stabilization via the Task Space Control (TSC) to drive $\mathcal{Y}_2\rightarrow 0 ,  \mathcal{Y}_1\rightarrow 0$. Typically, the joint accelerations, motor torques, and ground reaction forces are all included as optimization variables in the TSC formulation. Here we present the TSC formulation with only the motor torques as the optimization variables, which lets the TSC based quadratic program be solved efficiently.

First, we notice that the ground reaction forces are affine functions of the input torques $u$ as:
\begin{align}
 F_v &= A_v u + b_v, \label{utoF}\\
 A_{v} &=  - (J_v M^{-1} J_v^{T} )^{-1} J_v M^{-1} B,\\
 b_{v} &= (J_v M^{-1} J_v^{T} )^{-1} (J_v M^{-1}H -\dot{J}_v \dot{q} ).
\end{align}
Plugging this into the equation of motion yields,
\begin{equation}
 M(q)\ddot{q} + \underset{\bar{H}}{\underbrace{ H(q,\dot{q}) - J_s^T \tau_s(q)- J_{h,v}^T b_v}} =\underset{\bar{B}_v}{\underbrace{ (B  +  J_{h,v}^T A_{v}) }} u , \nonumber
\end{equation}
which yields the affine relation $\ddot{q}  = \bar{A}u + \bar{b}$ with
\begin{equation} \bar{A}= M^{-1} \bar{B}_v, \bar{b} = -M^{-1} \bar{H}.\end{equation}
As the derivatives of outputs are affine w.r.t. $\ddot{q}$, they are also affine w.r.t to $u$:
\begin{align}
\dot{\mathcal{Y}_1} &= A_1 \ddot{q} + b_1 = \mathcal{A}_1 u + \mathcal{B}_1,\\
\ddot{\mathcal{Y}_2} &= A_2 \ddot{q} + b_2 = \mathcal{A}_2 u + \mathcal{B}_2,
\end{align}
where the explicit expression of each term is omitted. Compactly,
\begin{equation}
    \dot{\mathcal{\mathbf{Y}}}  = \mathcal{A}u + \mathcal{B}
\end{equation} with $\mathcal{\mathbf{Y}} = [\mathcal{Y}_1; \dot{\mathcal{Y}}_2].$
The desired derivatives (accelerations) for stabilization are,
\begin{align}
\dot{\mathcal{Y}}^{\text{des}}_1 & = -K^p_1 \mathcal{Y}_1,  \\ \ddot{\mathcal{Y}}^{\text{des}}_2 &= -K^p_2 \mathcal{Y}_2 - K^d_2 \dot{\mathcal{Y}}_2.
\end{align}
This motivates a quadratic cost to minimize the difference between the actual and desired accelerations:
\begin{equation}
J_{\textrm{TSC}} (u) = (\mathcal{A}u + \mathcal{B} - \dot{\mathcal{Y}}^{des})^T(\mathcal{A}u + \mathcal{B} - \dot{\mathcal{Y}}^{des}),
\end{equation}
with $\dot{\mathcal{Y}}^{des} = [\dot{\mathcal{Y}_1}^{{\text{des}}^T}, \ddot{\mathcal{Y}_2}^{{\text{des}}^T}]^T$.

Additionally, two physical constraints are included for the feasibility of the torques on the robots. In the ground phase, the ground reaction forces have to satisfy the physics constraints such as non-negative normal forces and non-slipping condition, formulated by $C_v F_v \leq 0$, where $C_v$ is a constant matrix. The constraint is also on $u$ from Eq. \eqref{utoF}:
\begin{equation}
 C_v A_v u \leq -C_v b_v.
\end{equation}
Moreover, the motor torque must be within the feasible limits of the robot hardware. For electrical motors, the available motor torque depends on the rotational speed. Then we encode that motor torque limit constraint as:
\begin{equation}
\label{power}
    u_{lb}(\dot{q}) \leq u \leq u_{ub}(\dot{q}).
\end{equation}
Thus the quadratic program based controller for each domain is formulated as:
\begin{align}
\label{QP}
\quad u^{*} =  \underset{u \in \mathbb{R}^{11}} {\text{argmin}}&  \quad J_{\textrm{TSC}} (u), \nonumber \\
\text{s.t.}  \quad &  C_v A_v u \leq -C_v b_v,   \ \  \tag{GRF}    \nonumber  \\
 \quad  & u_{lb}(\dot{q}) \leq u \leq u_{ub}(\dot{q}).  \nonumber \tag{Torque Limit}
 \end{align}

\section{Results and Discussion}
\label{sec:results}
The proposed planning and optimization-based controller are implemented in simulation on the modified robot model of Cassie. The dynamics are integrated using Matlab ODE45 with event triggering. The trajectory optimization is solved via IPOPT \cite{wachter2006implementation}. Due to the simple dynamics, the optimizations are typically solved within seconds without code optimization for speed. The TSC is solved at 2kHz using qpOASES \cite{Ferreau2014}. The results can be seen in \cite{Results}. 

Using the proposed method, we generated several somersault motions on the robot. We specify the desired jumping height and forward distance, and then the optimization on Flywheel SLIP is performed. Jumping is thus generated on the robot. When the robot lifts off, the momentum transmission control is applied. After touch-down, the landing optimization is performed. Then the robot lands to a static configuration by tracking the optimized landing trajectory. 

Even though the model difference between the Flywheel SLIP and the robot can cause about 0.1m difference on the jumping height, the sequential planning and control still can achieve successful landing and accurate forward distance. Due to the numerical precision of integration, the robot can behave asymmetrically across the frontal place; since lateral COM position is controlled at the ground phase, the robot is still stabilized. Fig. \ref{results} demonstrates four somersaults. Fig. \ref{results2} show the trajectories of the produced somersault A.

\textbf{Robust to ground height change:} Suppose the ground height of the landing is not exactly known, the robot can still land successfully. The reason is that landing planning is performed after the post-impact event. The somersault D in Fig. \ref{results} is an example where the ground height increases by 0.1m, which shows the robustness of the planning and control with respect to the ground height change. 

\textbf{Practical Feasibility:} The result in Fig. \ref{results2} is on the robot with flywheel I selected in Table I, which has relatively small inertia comparing to the inertia of the rest of the robot. Therefore, the required rotational speed is high, and we replace Eq. \eqref{power} on the flywheel by a constant torque constraint to explore the required actuation. As a comparison, the knee motor module of Cassie has a maximum torque of $195.2$(Nm); while the maximum speed after gear reduction is only $81.25$(rpm). This implies that we need a significant improvement on the actuation design on the flywheel to realize such somersaulting motion. 

We also tried different flywheel specifications in Table I to realize somersault C in Fig. \ref{results}, with an apex height of $1.2$m from the platform. By limiting the maximum joint torque, the maximum required flywheel speed decreases with the increases of the flywheel inertia. However, if the mass increases, the lower body requires an increase of the maximum torques to generate the desired apex height. If the desired apex increases, which leads to a lower required flywheel momentum due to a longer flight duration, the knee actuation has to be increased. Another way to decrease the required momentum generation on the flywheel is to increase the hip pitch rotational speed at lift-off. We found that the contribution to decreasing the flywheel momentum is small unless the maximum hip joint speed is significantly increased. With the above discussion, we conclude that the practical realization of somersaults relies on powerful actuation design and good inertia distribution. This is a challenging topic to study in the field of legged locomotion. 

\begin{table}[b]
\caption{Different Flywheel specifications } 
\label{ModelTable}
\centering
\begin{tabular} {|c||c|c|c|c|}
\hline
Flywheel & $I_{\text{pitch}}$ ($\text{kg} \cdot \text{m}^2$) & $m$ (kg) & $\tau_{\text{max}}$ (Nm)  & $w_{\text{max}}$ (rpm) \\
\hline 
\hline
I  & $0.1074$ & $9.54$ & $195$ & $1645$ \\
\hline
II  & $0.2148$ & $19.09$ & $195$ & $812$ \\
\hline
III  & $0.3222$ & $28.63$ & $195$ & $573$ \\
\hline
\end{tabular}
\end{table}



\section{Conclusion and Future Work}
To conclude, we proposed a sequential planning and control method for generating somersaults on bipedal robots. The planning on the ground phase is based on the Flywheel SLIP model, and that in the flight phase follows the principle of conservation of momentum using momentum transmission. The planned trajectories are extracted and tracked by momentum-included task space control on the robot Cassie.

Future work will be devoted to understanding the necessary components in achieving other types of aerial motions, including periodic hopping, running, and aerial flipping in the transversal and coronal planes. Additionally, it is interesting to study how to generate somersaults on robots without free-rotational joints. From our perspective, the bottleneck for achieving this kind of highly dynamic motion on the physical robot is on the necessity to unify mechanical design with control implementation at the system level. Future work will explore this trade-off with a view toward control-informed robot design for realizing dynamic aerial motion. 
\addtolength{\textheight}{-0cm}
\bibliographystyle{IEEEtran}
\bibliography{main}
\end{document}